\documentclass{ieeeaccess}
\usepackage{cite}
\usepackage{amsmath,amssymb,amsfonts}
\usepackage{algorithmic}
\usepackage{graphicx}
\usepackage{textcomp}
\usepackage{xcolor}
\usepackage{makecell}
\usepackage{hyperref}
\usepackage{amsmath}
\usepackage{amssymb}
\usepackage[ruled,vlined,noend]{algorithm2e}

\newcommand\hl[1]{#1}

\def\BibTeX{{\rm B\kern-.05em{\sc i\kern-.025em b}\kern-.08em
    T\kern-.1667em\lower.7ex\hbox{E}\kern-.125emX}}
\begin{document}
\history{Manuscript accepted to IEEE Access}
\doi{10.1109/ACCESS.2021.3118207}

\title{EvoPose2D: Pushing the Boundaries of 2D Human Pose Estimation using Accelerated Neuroevolution with Weight Transfer}
\author{
\uppercase{William McNally}, 
\uppercase{Kanav Vats},
\uppercase{Alexander Wong}, \IEEEmembership{Senior Member, IEEE},
\uppercase{and John McPhee}
}
\address[]{Department of Systems Design Engineering, University of Waterloo, Waterloo, ON N2L 3G1}
\address[]{Waterloo Artificial Intelligence Institute, University of Waterloo, Waterloo, ON, N2L 3G1}
\tfootnote{This work was supported financially by the Canada Research Chairs Program, the Natural Sciences and Engineering Research Council of Canada, and a Google Cloud Academic Research Grant. We also acknowledge Compute Canada, an NVIDIA GPU Grant, and the TPU Research Cloud Program for hardware support. 
}

\markboth
{McNally \headeretal: Preparation of Papers for IEEE TRANSACTIONS and JOURNALS}
{McNally \headeretal: Preparation of Papers for IEEE TRANSACTIONS and JOURNALS}

\corresp{Corresponding author: William McNally (e-mail: wmcnally@uwaterloo.ca).}

\begin{abstract}
Neural architecture search has proven to be highly effective in the design of efficient convolutional neural networks that are better suited for mobile deployment than hand-designed networks. Hypothesizing that neural architecture search holds great potential for human pose estimation, we explore the application of neuroevolution, a form of neural architecture search inspired by biological evolution, in the design of 2D human pose networks for the first time. Additionally, we propose a new weight transfer scheme that enables us to accelerate neuroevolution in a flexible manner. Our method produces network designs that are more efficient and more accurate than state-of-the-art hand-designed networks. In fact, the generated networks process images at higher resolutions using less computation than previous hand-designed networks at lower resolutions, allowing us to push the boundaries of 2D human pose estimation. Our base network designed via neuroevolution, which we refer to as EvoPose2D-S, achieves comparable accuracy to SimpleBaseline while being 50\% faster and 12.7x smaller in terms of file size. Our largest network, EvoPose2D-L, achieves new state-of-the-art accuracy on the Microsoft COCO Keypoints benchmark, is 4.3x smaller than its nearest competitor, and has similar inference speed. The code is publicly available at \url{https://github.com/wmcnally/evopose2d}.
\end{abstract}


\titlepgskip=-15pt

\maketitle

\section{Introduction}
Two-dimensional human pose estimation is a visual recognition task dealing with the autonomous localization of anatomical human joints, or more broadly, ``keypoints,'' in RGB images and video~\cite{toshev2014deeppose, tompson2014joint, andriluka20142d, kim2020pose, wang2019human}. It is widely considered a fundamental problem in computer vision due to its many downstream applications, including action recognition~\cite{cheron2015p, ullah2017action, el2018human, mcnally2019star, mcnally2018action, mcnally2019golfdb} and human tracking~\cite{insafutdinov2017arttrack, andriluka2018posetrack, xiao2018simple}. In particular, it is a precursor to 3D human pose estimation~\cite{martinez2017simple, pavllo20193d, liang2020adaptive}, which serves as a potential alternative to invasive marker-based motion capture.

In line with other streams of computer vision, the use of deep learning~\cite{chen2014big, lecun2015deep}, and specifically deep convolutional neural networks~\cite{lecun1995convolutional} (CNNs), has been prevalent in 2D human pose estimation~\cite{toshev2014deeppose, tompson2014joint, newell2016stacked, cao2017realtime, chen2018cascaded, xiao2018simple, sun2019deep}. The most accurate 2D human pose estimation methods use a two-stage, top-down pipeline, where an off-the-shelf person detector is first used to detect human instances in an image, and the 2D human pose network is run over the person detections to obtain keypoint predictions~\cite{chen2018cascaded, xiao2018simple, sun2019deep}. This paper focuses on the latter stage of this commonly used top-down pipeline, but we emphasize that our method is applicable to the design of bottom-up human pose estimation networks~\cite{cao2017realtime, cheng2020higherhrnet} as well.

Recently, there has been a growing interest in the use of machines to help design CNN architectures through a process known as neural architecture search (NAS)~\cite{zoph2017neural, baker2016designing, wistuba2019survey, weng2019unet}. NAS removes human bias from the design process and permits the automated exploration of diverse network architectures that often transcend human intuition and provide greater accuracy using less computation. Moreover, networks designed using NAS often have fewer parameters~\cite{tan2019mnasnet}, which reduces the need for expensive main memory access on embedded hardware designed with small memory caches~\cite{sandler2018mobilenetv2}. Despite the widespread success of NAS in many areas of computer vision~\cite{tan2019efficientnet, chen2018searching, liu2019auto, ghiasi2019fpn, nekrasov2019fast, zhu2019v, yan2021lighttrack}, the design of 2D human pose networks has remained, for the most part, human-principled. 

\begin{figure}
\centering
    \includegraphics[trim={2mm, 2mm, 0, 0}, clip, width=1.0\linewidth]{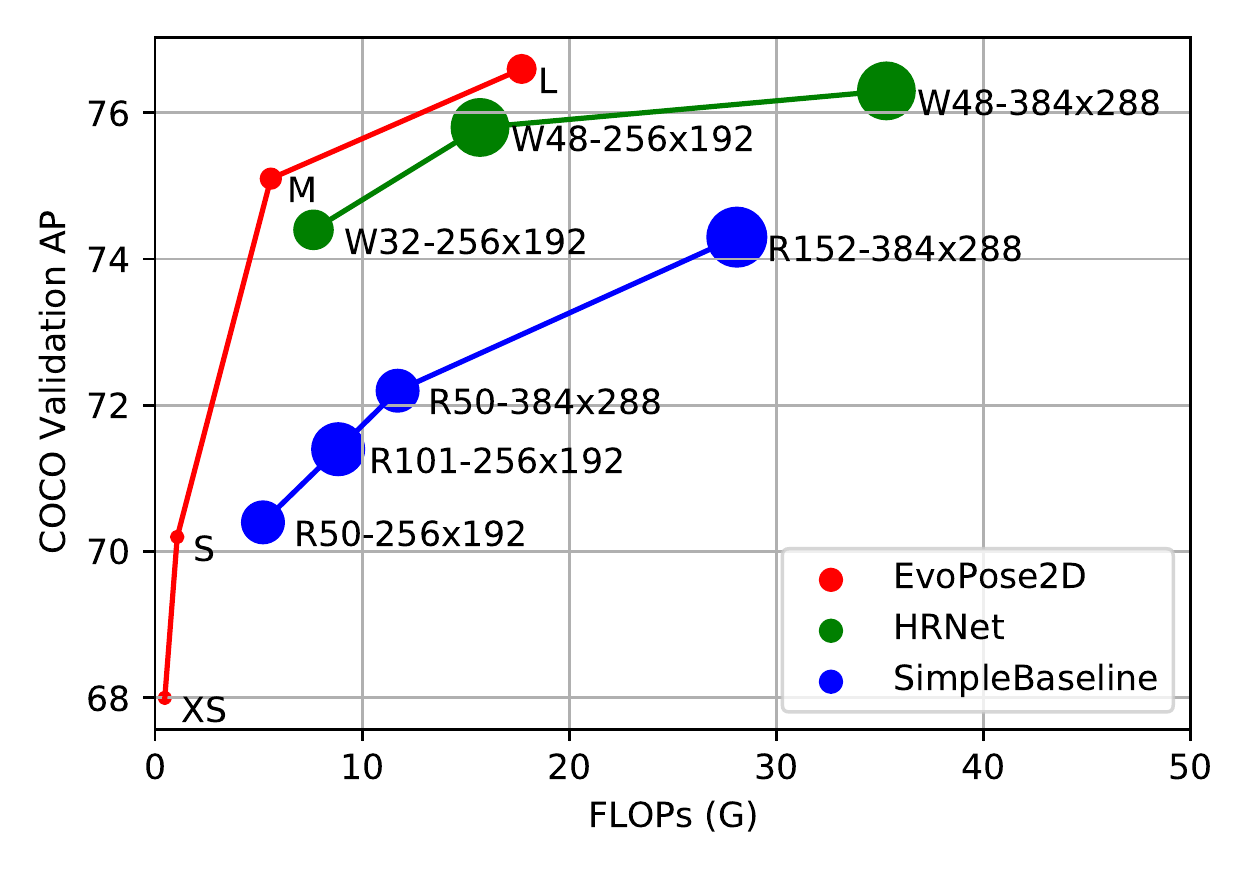}
\caption{A comparison of the accuracy, size, and computational cost of EvoPose2D, SimpleBaseline~\cite{xiao2018simple}, and HRNet\cite{sun2019deep} at different scales. The circle size is proportional to the network file size. EvoPose2D-S provides comparable accuracy to SimpleBaseline (ResNet-50), is 12.7x smaller, and uses 4.9x fewer FLOPs. At full-scale, EvoPose2D-L obtains state-of-the-art accuracy using 2.0x fewer FLOPs and is 4.3x smaller compared to HRNet-W48. In contrast to the referenced methods, our EvoPose2D results do not make use of model-agnostic enhancements such as ImageNet pretraining, half-body augmentation, or non-maximum suppression during post-processing.}
\label{fig:teaser}
\end{figure}

In this study, we explore the application of neuroevolution~\cite{rodrigues2020study}, a realization of NAS inspired by evolution in nature, to 2D human pose estimation for the first time. To run large-scale NAS experiments within a practical time-frame, we propose a new weight transfer scheme that is highly flexible and accelerates neuroevolution. We exploit this weight transfer scheme, along with large-batch training on high-bandwidth Tensor Processing Units (TPUs), to run fast neuroevolutions within a search space geared towards 2D human pose estimation. Our neuroevolution framework produces a 2D human pose network that has a relatively simple design, provides state-of-the-art accuracy when scaled, and uses fewer floating-point operations (FLOPs) and parameters than the best performing networks in the literature (see Fig.\ \ref{fig:teaser}). The key contributions of this research are summarized as follows:

\begin{itemize}
    \item We propose a new weight transfer scheme to accelerate neuroevolution and apply neuroevolution to 2D human pose estimation for the first time. In contrast to previous neuroevolution methods that exploit weight transfer, our method is not constrained by complete function preservation~\cite{wistuba2018deep, wei2016network}. Despite relaxing this constraint, our experiments indicate that the level of functional preservation afforded by our weight transfer scheme is sufficient to provide fitness convergence, thereby simplifying neuroevolution and making it more flexible.
    \item We present empirical evidence that large-batch training (i.e., batch size of 2048) can be used in conjunction with the Adam optimizer~\cite{kingma2014adam} to accelerate the training of 2D human pose networks with no loss in accuracy. We reap the benefits of large-batch training in our neuroevolution experiments by maximizing training throughput on high-bandwidth TPUs.
    \item We design a search space conducive to 2D human pose estimation and leverage the above contributions to run a fast full-scale neuroevolution of 2D human pose networks ($\sim$1 day using eight v2-8 TPUs). As a result, we are able to produce a computationally efficient 2D human pose estimation model that achieves state-of-the-art accuracy on the most widely used benchmark dataset. 
\end{itemize}

\section{Related Work}
This work draws upon several areas of deep learning research to engineer a high-performing 2D human pose estimation model. We review the three most relevant areas of the literature in the following sections.


\subsection{Large-batch Training of Deep Neural Networks}
It has been shown that training deep neural networks using large batch sizes with stochastic gradient descent causes a degradation in the quality of the model as measured by its ability to generalize to unseen data~\cite{hoffer2017train, keskar2016large}. Recently, Goyal et al.\ \cite{goyal2017accurate} implemented measures for mitigating the training difficulties caused by large batch sizes, including linear scaling of the learning rate, and an initial warmup period where the learning rate is gradually increased. 

Maximizing training efficiency using large-batch training is critical when the computational demand of training is very high, such as in neural architecture search. However, deep learning methods are often data-dependent and task-dependent, so it remains unclear whether the training measures imposed by Goyal et al.\ for image classification apply in the general case. It is also unclear whether the learning rate modifications are applicable to optimizers that use adaptive learning rates. Adam~\cite{kingma2014adam} is an example of such an optimizer and is widely used in 2D human pose estimation. In this paper, we empirically investigate the use of large batch sizes in conjunction with the Adam optimizer in the training of 2D human pose networks.




\subsection{2D Human Pose Estimation using Deep Learning} The first use of deep learning for human pose estimation came in 2014, when Toshev and Svegedy~\cite{toshev2014deeppose} regressed 2D keypoint coordinates directly from RGB images using a cascade of deep CNNs. Arguing that the direct regression of pose vectors from images was a highly non-linear and difficult to learn mapping, Tompson et al.\ \cite{tompson2014joint} introduced the notion of learning a \textit{heatmap} representation. Mean squared error (MSE) was used to minimize the distance between the predicted and target heatmaps, where the targets were generated using Gaussians with small variance centered on the ground-truth keypoint coordinates. 




Several of the methods that followed built upon iterative heatmap refinement in a multi-stage fashion including intermediate supervision~\cite{wei2016convolutional, cao2017realtime, newell2016stacked}. Remarking the inefficiencies associated with multi-stage stacking, Chen et al.~\cite{chen2018cascaded} proposed the Cascaded Pyramid Network (CPN), a holistic network constructed using a ResNet-50~\cite{he2016deep} feature pyramid~\cite{lin2017feature}. Xiao et al.\ \cite{xiao2018simple} presented yet another single-stage architecture called SimpleBaseline, which stacked transpose convolutions on top of ResNet. Sun et al.\ \cite{sun2019deep} demonstrated with HRNet that maintaining high-resolution features throughout the entire network could provide greater accuracy. HRNet represents the state-of-the-art in 2D human pose estimation among peer-reviewed works at the time of writing. 


An issue surrounding the 2D human pose estimation literature is that it is often difficult to make fair comparisons of model performance due to the heavy use of model-agnostic improvements. Examples include the use of different learning rate schedules~\cite{sun2019deep, li2019rethinking}, more data augmentation~\cite{li2019rethinking, bin2020adversarial}, loss functions that target more challenging keypoints~\cite{chen2018cascaded}, specialized post-processing steps~\cite{moon2019posefix, huang2020devil}, or more accurate person detectors~\cite{li2019rethinking, huang2020devil}. These discrepancies in training algorithms can potentially account for the reported differences in accuracy. To directly compare our pose estimation architectures with the state-of-the-art, we re-implement SimpleBaseline~\cite{xiao2018simple} and HRNet~\cite{sun2019deep} and train all networks under the same settings using the same hardware.

\subsection{Neuroevolution}
Neuroevolution is a form of neural architecture search that harnesses evolutionary algorithms to search for optimal network architectures~\cite{real2017large, stanley2019designing}. Network morphisms~\cite{wei2016network} and function-preserving mutations~\cite{wistuba2018deep} are techniques that reduce the computational cost of neuroevolution. In essence, these methods iteratively mutate networks and perform weight transfer in such a way that the function of the network is completely preserved upon mutation, i.e., the output of the mutated network is identical to that of the parent network. Ergo, the mutated child networks need only be trained for a relatively small number of steps compared to when training from a randomly initialized state. As a result, these techniques are capable of reducing the search time to a matter of GPU days. However, function-preserving mutations can be challenging to implement and also restricting (e.g., complexity cannot be reduced~\cite{wistuba2018deep}). Our proposed weight transfer scheme serves as a more flexible alternative that addresses these issues, is effective in accelerating neuroevolution, and has a simpler implementation.

We briefly discuss the important distinctions between neuroevolution methods that leverage weight \textit{transfer}, and alternative NAS approaches that leverage weight \textit{sharing}, such as ENAS~\cite{pham2018efficient} and DARTS~\cite{liu2018darts}. Weight sharing approaches are sometimes referred to as one-shot architecture search~\cite{bender2018understanding}, because architectures are sampled from a single, over-parameterized supergraph encompassing the entire search space (one-shot model). The search is performed over a single training run of the supergraph, where subgraphs are selected, evaluated using the supergraph weights, and then ranked. The best performing subgraph is finally trained from scratch. One-shot methods are based around the hypothesis that the ranking of the candidate subgraphs correlates with their true ranking following final training. However, Yu et al.\ observe that this correlation is very weak, and ultimately find that ENAS and DARTS perform no better than a random search~\cite{yu2019evaluating}. Moreover, some one-shot methods require the entire supergraph to be kept in memory, which inherently limits the size of the search space. These issues are not a concern in neuroevolution because the candidate architectures are trained separately and thus do not share weights. In a recent benchmarking of NAS algorithms, neuroevolution methods were among the top performing algorithms and consistently outperformed random search~\cite{ying2019bench}.

NAS algorithms have predominantly been developed and evaluated on small-scale image datasets~\cite{wistuba2019survey}. The use of NAS in more complex visual recognition tasks remains limited, in large part because the computational demands make it impractical. This is especially true for 2D human pose estimation, where training a single model can take several days~\cite{chen2018cascaded}. Nevertheless, the use of NAS in the design of 2D human pose networks has been attempted in a few cases~\cite{yang2019pose, gong2020autopose, zhang2020cpnas}. Although some of the resulting networks provided superior computational efficiency as a result of having fewer parameters and operations, none managed to surpass the best performing hand-crafted networks in terms of accuracy. 




\section{Accelerating Neuroevolution using Weight Transfer}
\label{sec:weight_transfer}
Suppose that a \hl{pretrained ``parent'' neural} network is represented by the function $\mathcal{P}\left(\mathbf{x}\,|\,\mathbf{\theta}^\mathcal{(P)}\right)$, where $\mathbf{x}$ is the input to the network and $\mathbf{\theta}^\mathcal{(P)}$ are its \hl{learned} parameters. The foundation of the proposed neuroevolution framework lies in the process by which the \hl{unknown} parameters $\mathbf{\theta}^\mathcal{(C)}$ in a mutated child network $\mathcal{C}$ are inherited from $\mathbf{\theta}^\mathcal{(P)}$ such that $\mathcal{C}\left(\mathbf{x}\,|\,\mathbf{\theta}^\mathcal{(C)}\right) \approx \mathcal{P}\left(\mathbf{x}\,|\,\mathbf{\theta}^\mathcal{(P)}\right)$. That is, the output, or ``function,'' of the mutated child network is similar to the parent, but not necessarily equal. To enable fast neural architecture search, the degree to which the parent's function is preserved must be sufficient to allow $\mathbf{\theta}^\mathcal{(C)}$ to be trained to convergence in a small fraction of the number of steps normally required when training from a randomly initialized state. 

\begin{figure}
\centering
    \includegraphics[width=1.0\linewidth]{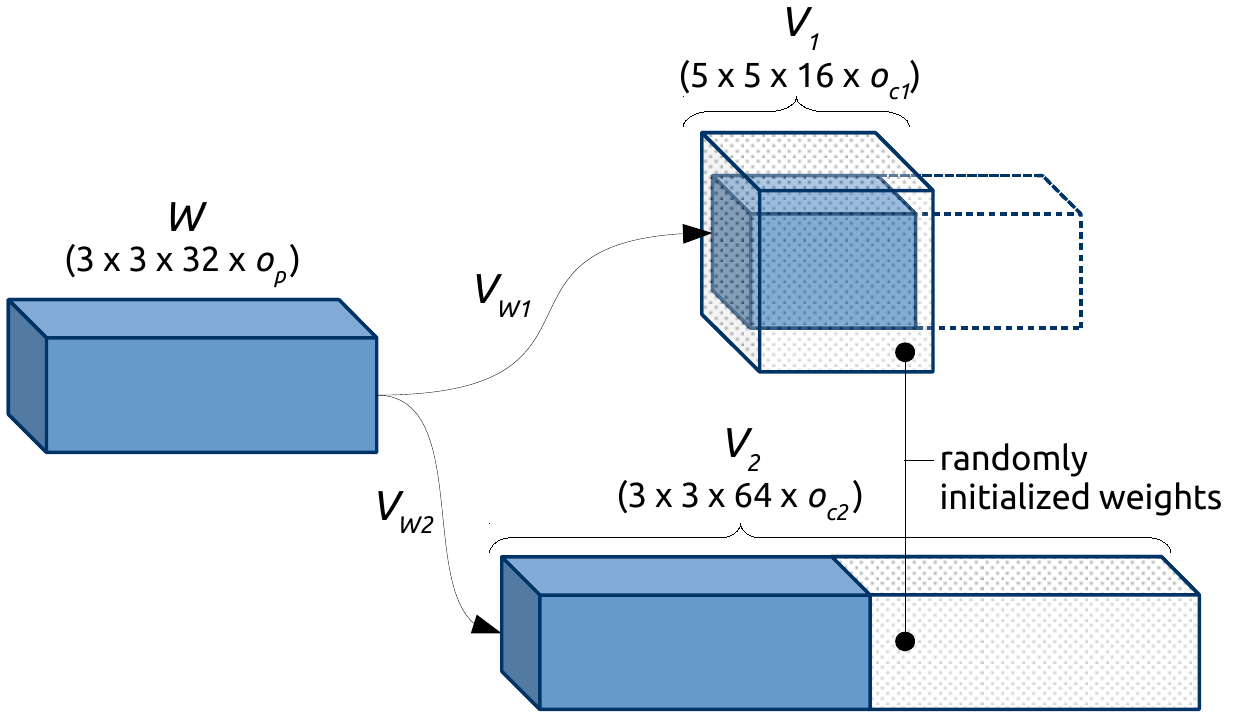}
\caption{Two examples ($W\rightarrow V_1$, $W\rightarrow V_2$) of the weight transfer used in the proposed neuroevolution framework. The trained weights (shown in blue) in the parent convolutional filter $W$ are transferred, either in part or in full ($V_W$), to the corresponding filter $V$ in the mutated child network. The weight transfer extends to all output channels in the same manner as depicted here for input channels.}
\label{fig:weight_transfer}
\end{figure}

To formalize the proposed weight transfer in the context of 2D convolution, we denote $W^{(l)} \in \mathbb{R}^{k_{p1} \times k_{p2} \times i_p \times o_p}$ as the weights used by layer $l$ of the parent network, and $V^{(l)} \in \mathbb{R}^{k_{c1} \times k_{c2} \times i_c \times o_c}$ as the weights of the corresponding layer in the mutated child network, where $k$ is the kernel size, $i$ is the number of input channels, and $o$ is the number of output channels. For the sake of brevity, we consider the special case when $k_{p1} = k_{p2} = k_p$, $k_{c1} = k_{c2} = k_c$, and $o_p = o_c$, but the following definition can easily be extended to when $k_{p1} \neq k_{p2}$, $k_{c1} \neq k_{c2}$, or $o_p \neq o_c$. The inherited weights $V_W$ are given by:
\begin{equation*}
V_W^{(l)} = 
\begin{cases} 
      W^{(l)}_{p:p+k_c, \, p:p+k_c, \, :i_c, \, :} & (i_c < i_p) \land (k_c < k_p) \medskip\\
      W^{(l)}_{p:p+k_c, \, p:p+k_c, \, :, \,:} & (i_c \geq i_p) \land (k_c < k_p) \medskip\\
      W^{(l)}_{:, \, :, \, :i_c, \,:} & (i_c < i_p) \land (k_c \geq k_p) \medskip\\
      W^{(l)} & (i_c \geq i_p) \land (k_c \geq k_p) \medskip
\end{cases}
\end{equation*}
where $p = \frac{1}{2}(k_p-k_c)$. $V_W$ is transferred to $V$ and the remaining non-inherited weights in $V$ are randomly initialized. An illustration of the weight transfer between two convolutional layers is shown in Fig.\ \ref{fig:weight_transfer}. In principle, the proposed weight transfer can be used with convolutions of any dimensionality (e.g., 1D, 2D, or 3D convolutions), and is permitted between convolutional operators with different kernel size, stride, dilation, input channels, and output channels.  More generally, it can be applied to any operations with learnable parameters, including batch normalization and dense layers.

In essence, the proposed weight transfer method relaxes the function-preservation constraint imposed in~\cite{wei2016network, wistuba2018deep}. In practice, we find that the proposed weight transfer preserves the majority of the function of deep CNNs following mutation. This enables us to perform network mutations in a simple and flexible manner while maintaining good parameter initialization in the mutated network. As a result, the mutated networks can be trained using fewer iterations, which accelerates the neuroevolution.

\section{Fast Neuroevolution of 2D Human Pose Networks}
\label{sec:method}
This section includes the engineering details for our neuroevolution implementation that leverages the proposed weight transfer scheme to accelerate the evolution of a 2D human pose network. While we focus on the application of 2D human pose estimation, we note that our neuroevolution approach is generally applicable to all types of deep networks.

\subsection{Search Space}
Neural architecture search helps moderate human involvement in the design of deep neural networks. However, neural architecture search is by no means fully automatic. To some extent, our role transitions from a network designer to a search designer. Decisions regarding the search space are particularly important because the search space encompasses all possible solutions to the optimization problem, and its size correlates with the amount of computation required to thoroughly explore the space. As such, it is common to exploit prior knowledge in order to reduce the size of the search space and ensure that the sampled architectures are tailored toward the task at hand~\cite{zoph2018learning}. 

Motivated by the simplicity and elegance of the SimpleBaseline architecture~\cite{xiao2018simple}, we search for an optimal human pose estimation backbone using a search space inspired by~\cite{tan2019mnasnet, tan2019efficientnet}. Specifically, the search space encompasses a single-branch hierarchical structure that includes seven modules stacked in series. Each module is constructed of chain-linked inverted residual blocks~\cite{sandler2018mobilenetv2} that use an expansion ratio of six and squeeze-excitation~\cite{hu2018squeeze}. For each module, we search for the optimal kernel size, number of inverted residual blocks, and output channels. Considering the newfound importance of spatial resolution in the deeper layers of 2D human pose networks~\cite{sun2019deep}, we additionally search for the optimal stride of the last three modules. Without going into too much detail, our search space can produce $10^{14}$ unique backbones. To complete the network, an initial convolutional layer with 32 output channels precedes the seven modules, and three transpose convolutions with kernel size of 3x3, stride of 2, and 128 output channels are used to construct the network head. A diagram of the search space is provided in Fig.~\ref{fig:search_space}. Additional search space details are provided in Appendix~\ref{sec:app_search_space}.

\begin{figure}[h]
\centering
\includegraphics[trim={0, 9.4cm, 15cm, 0}, clip, width=1\linewidth]{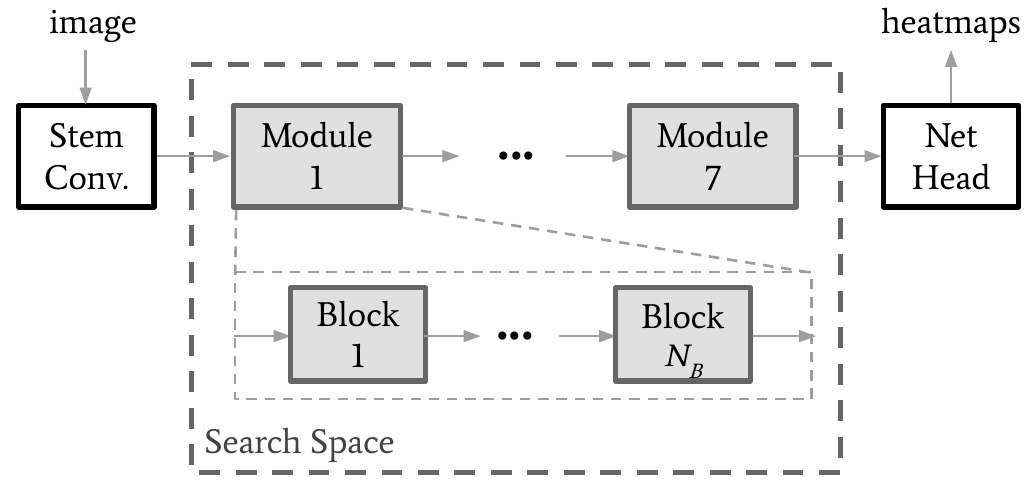}
\caption{Search space diagram.}
\label{fig:search_space}
\end{figure}

\subsection{Fitness} 
To strike a balance between computational efficiency and accuracy, we perform a multi-objective optimization that minimizes a fitness function including the validation loss and the number of network parameters. Given a 2D pose network represented by the function $\mathcal{N}\left(\mathbf{x}\,|\,\mathbf{\theta}^\mathcal{(\mathcal{N})}\right)$, the loss \hl{$\mathcal{L}_i$} for a single RGB input image $\textbf{I}\in\mathbb{R}^{h \times w \times 3}$ and corresponding target heatmap $\mathbf{S}\in\mathbb{R}^{h' \times w' \times K}$ is given by

\newcommand{\norm}[1]{\left\lVert#1\right\rVert}
\begin{equation}
\label{eq:loss}
    \mathcal{L}_i(\mathcal{N, \mathbf{I}}) = \frac{1}{K}\sum_{j=1}^{K} \delta(v_j > 0) \norm{\mathcal{N}\left(\mathbf{I}\,|\,\mathbf{\theta}^{(\mathcal{N})}\right)_j -  \textbf{S}_j}^2_2
\end{equation}
where $K$ is the number of keypoints and $v$ represents the keypoint visibility flags\footnotemark[1]. The target heatmaps $\mathbf{S}$ are generated by centering 2D Gaussians with a standard deviation of $\frac{h'}{64}$ pixels on the ground-truth keypoint coordinates \hl{and normalizing to a maximum intensity of 255. The overall validation loss is computed as:}
\begin{equation}
\label{eq:val_loss}
    \mathcal{L}(\mathcal{N}) = \frac{1}{N}\sum_{i=1}^{N}\mathcal{L}_i(\mathcal{N}, \mathbf{I}_i) 
\end{equation}
where $N$ is the number of image samples in the validation dataset.
Finally, the fitness of a network $\mathcal{N}$ is given by:
\begin{equation}
    \mathcal{J}(\mathcal{N}) = \left(\frac{T}{n(\theta^{\mathcal{N}})}\right)^\Gamma \mathcal{L}(\mathcal{N})
\end{equation}
where $n(\theta^{\mathcal{N}})$ is the number of parameters in $\mathcal{N}$, $T$ is the target number of parameters, and $\Gamma$ controls the fitness trade-off between the number of parameters and the validation loss. Minimizing the number of parameters instead of the number of floating-point operations (FLOPs) allows us to indirectly minimize FLOPs while not penalizing mutations that decrease the stride of the network. 

\subsection{Evolutionary Strategy}
The evolutionary strategy proceeds as follows. In generation ``0'', a common ancestor network is manually defined and trained from scratch for $e_0$ epochs. In generation 1, $\lambda$ children are generated by mutating the ancestor network. The mutation details are provided in Appendix~\ref{sec:app_mutation}. The weight transfer outlined in Section~\ref{sec:weight_transfer} is performed between the ancestor and each child (additional implementation details provided in Appendix~\ref{sec:app_wt}), after which the children's weights are trained for $e$ epochs ($e \ll e_0$). At the end of generation 1, the $\mu$ networks with the best fitness from the pool of ($\lambda + 1$) networks (children + ancestor) become the parents in the next generation. In generation 2 and beyond, the mutation $\rightarrow$ weight transfer $\rightarrow$ training process is repeated and the top-$\mu$ networks from the pool of ($\lambda + \mu$) networks (children + parents) become the parents in the next generation. The evolution continues until manual termination, typically after the fitness has converged. 

\subsection{Large-batch training}
Even with the computational savings afforded by weight transfer, running a full-scale neuroevolution of 2D human pose networks at a standard input resolution of 256x192 would not be feasible within a practical time-frame using common GPU resources (e.g., 8-GPU server). To reduce the search time to within a practical range, we exploit large batch sizes when training 2D human pose networks on TPUs. In line with~\cite{goyal2017accurate}, we linearly scale the learning rate with the batch size and gradually ramp-up the learning rate during the first few epochs. In Section~\ref{sec:large_batch}, we empirically demonstrate that this training regimen can be used in conjunction with the Adam optimizer~\cite{kingma2014adam} to train 2D human pose networks up to a batch size of 2048 with no loss in accuracy. To our best knowledge, the largest batch size previously used to train a 2D human pose network was 256, which required 8 GPUs~\cite{li2019rethinking}.

\subsection{Compound scaling} 
It has been shown recently that scaling a network's resolution, width (channels), and depth (layers) together is more efficient than scaling one of these dimensions individually~\cite{tan2019efficientnet}. Motivated by this finding, we scale the base network found through neuroevolution to different input resolutions using the following depth ($c_d$) and width ($c_w$) coefficients:

\begin{equation}
\label{eq:scaling}
\begin{aligned}
c_d  = \alpha ^ \phi \quad c_w = \beta ^ \phi \quad \phi = \frac{\log r - \log r_s}{\log{\gamma}}
\end{aligned}
\end{equation}
where $r_s$ is the search resolution, $r$ is desired resolution, and $\alpha$, $\beta$, $\gamma$ are scaling parameters. For convenience, we use the same scaling parameters as in \cite{tan2019efficientnet} ($\alpha$ = 1.2, $\beta$ = 1.1, $\gamma$ = 1.15) but hypothesize that better results could be obtained if these parameters were tuned. 

\section{Experiments}
\vspace{4pt}
\subsection{Datasets}
\vspace{4pt}
\subsubsection{Microsoft COCO}
The 2017 Microsoft COCO Keypoints dataset~\cite{lin2014microsoft} is the predominant dataset used to evaluate 2D human pose estimation models. It contains over 200k images and 250k person instances labeled with 17 keypoints. We fit our models to the training subset, which contains 57k images and 150k person instances. We evaluate our models on both the validation and test-dev sets, which contain 5k and 20k images, respectively. We report the standard average precision (AP) and average recall (AR) scores based on Object Keypoint Similarity (OKS)\footnote{More details available at \href{https://cocodataset.org/\#keypoints-eval}{https://cocodataset.org/\#keypoints-eval}.}: AP (mean AP at OKS = 0.50, 0.55, $\dotsc$, 0.90, 0.95), AP$^{50}$ (AP at OKS = 0.50), AP$^{75}$, AP$^M$ (medium objects), AP$^L$ (large objects), and AR (mean AR at OKS = 0.50, 0.55, $\dotsc$, 0.90, 0.95).

\subsubsection{PoseTrack} 
PoseTrack~\cite{andriluka2018posetrack} is a large-scale benchmark for 2D human pose estimation and tracking in video. The dataset contains 1,356 video sequences, 46k annotated frames, and 276k person instances. The dataset was converted to COCO format and the COCO evaluation toolbox (\texttt{pycocotools}) was used to evaluate the accuracy in the multi-person human pose estimation task (i.e., using the same accuracy metrics as above). In experiments, we train our models on the 2018 training set (97k person instances), and evaluate on the 2018 validation set (45k person instances) using ground-truth bounding boxes.

\subsection{Large-batch training of 2D Human Pose Networks on TPUs}
\label{sec:large_batch}
To maximize training throughput on TPUs, we run experiments to investigate the training behaviour of 2D human pose networks using larger batch sizes than have been used previously. For these experiments, we re-implement the SimpleBaseline model of Xiao et al.\ \cite{xiao2018simple} and train it on the Microsoft COCO dataset. The SimpleBaseline network stacks three transpose convolutions with 256 channels and kernel size of 3x3 on top of a ResNet-50 backbone, which is pretrained on ImageNet~\cite{krizhevsky2012imagenet}. We run the experiments at an input resolution of $256\times192$, which yields output heatmap predictions of size $64\times48$. According to the TensorFlow profiler used, this model has 34.1M parameters and 5.21G FLOPs.

\subsubsection{Implementation Details}
\label{sec:tpu_details}
The following experimental setup was used to obtain the results for all models trained on COCO in this paper. Additional implementation details for neuroevolution and PoseTrack training are provided in Sections~\ref{sec:neuro_details} and~\ref{sec:sota}, respectively.
TensorFlow 2.3 and the \texttt{tf.keras} API were used for implementation. The COCO keypoints dataset was first converted to TFRecords for TPU compatibility (1024 examples per shard). The TFRecord dataset contains the serialized examples including the raw images, keypoint locations, and bounding boxes. The dataset was stored in a Google Cloud Storage Bucket where it was accessed remotely by the TPU host CPU over the network. Thus, all preprocessing, including target heatmap generation, image transformations, and data augmentation, was performed on the host CPU. A single-device v3-8 TPU (8 TPU cores, 16GB of high-bandwidth memory per core) was used for training, validation, and testing.


\medskip\noindent\textbf{Preprocessing.} The RGB input images were first normalized to a range of [0, 1], then centered and scaled by the ImageNet pixel means and standard deviations. The images were then transformed and cropped to the input size of the network. During training, random horizontal flipping, scaling, and rotation were used for data augmentation. The exact data augmentation configuration is provided in the linked code.

\medskip\noindent\textbf{Training.} The networks were trained for 200 epochs using bfloat16 floating-point format, which consumes half the memory compared to commonly used float32. The loss represented in Eq.\ (\ref{eq:val_loss}) was minimized using the Adam optimizer~\cite{kingma2014adam} with a cosine-decay learning rate schedule~\cite{loshchilov2016sgdr} and L2 regularization with $1\mathrm{e}{-5}$ weight decay. The base learning rate $l_r$ was set to $2.5\mathrm{e}{-4}$ and was scaled to $l_r\cdot\frac{n}{32}$, where $n$ is the global batch size. Additionally, a warmup period was implemented by gradually increasing the learning rate from $l_r$ to $l_r\cdot\frac{n}{32}$ over the first five epochs.  The validation loss was evaluated after every epoch using the ground-truth bounding boxes. 

\medskip\noindent\textbf{Testing.} The common two-stage, top-down pipeline was used during testing~\cite{chen2018cascaded, xiao2018simple, sun2019deep}. We use the same detections as~\cite{xiao2018simple, sun2019deep} and follow the standard testing protocol: the predicted heatmaps from the original and horizontally flipped images were averaged and the keypoint predictions were obtained after applying a quarter offset in the direction from the highest response to the second highest response. We do not use non-maximum suppression.

\begin{figure}[t]
\centering
    \includegraphics[trim={0, 2mm, 0, 0}, clip, width=\linewidth]{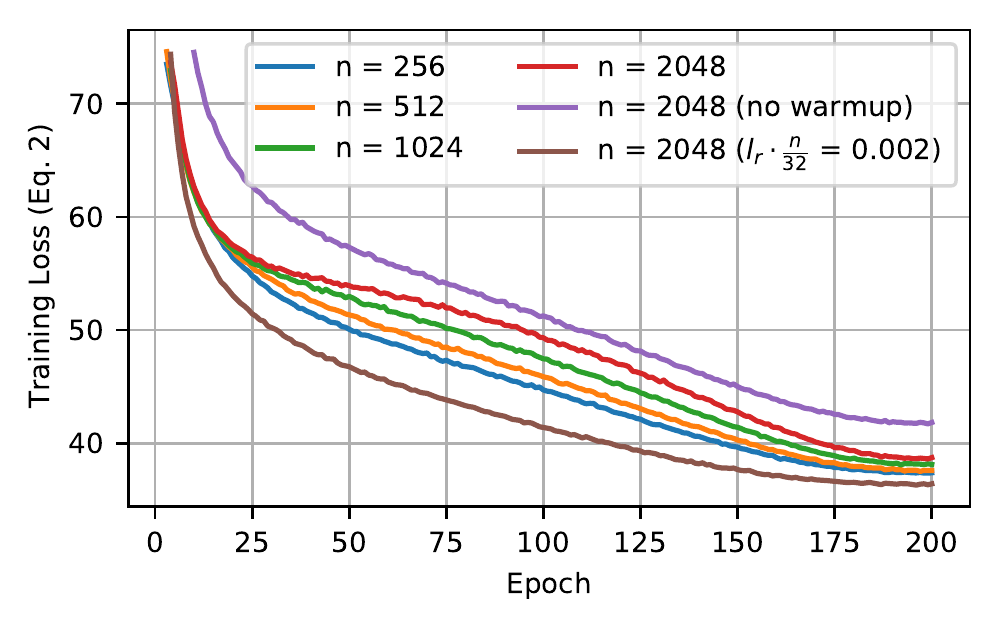}
    \includegraphics[trim={0, 2mm, 0, 2mm}, clip, width=\linewidth]{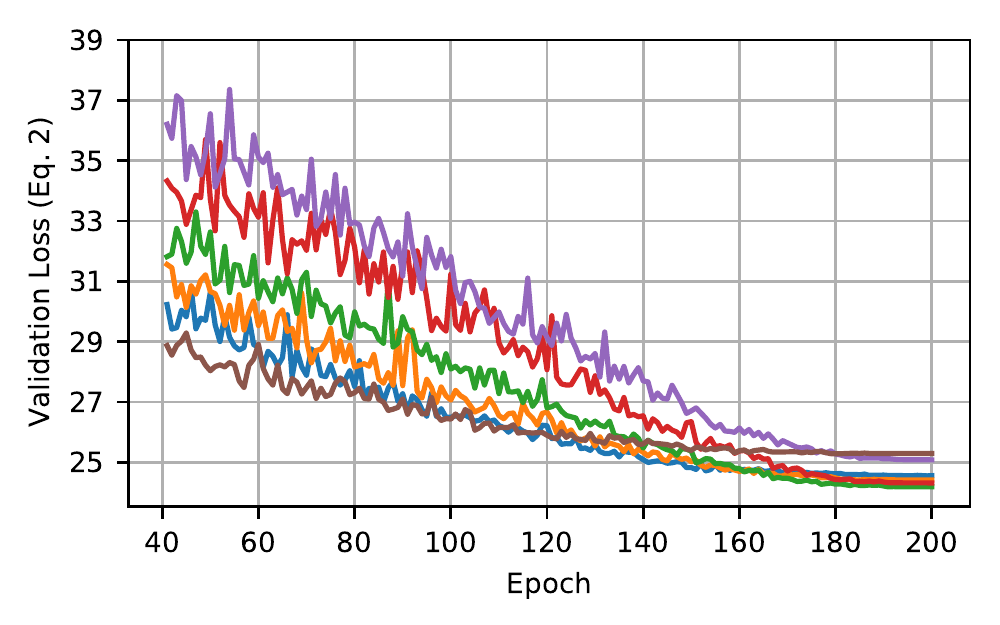}
\caption{Training (top) and validation (bottom) losses during training of SimpleBaseline (ResNet-50)~\cite{xiao2018simple} on a v3-8 Cloud TPU using various large batch sizes and learning rate schedules.}
\label{fig:sb_tpu}
\end{figure}

\subsubsection{Large Batch Training Results} 
\label{sec:sb_tpu}
The batch size was doubled from an initial batch size of 256 until the memory of the v3-8 TPU was exceeded. The maximum batch size attained was 2048. The loss curves for the corresponding training runs are shown in Fig.\ \ref{fig:sb_tpu}. While the final training loss increased marginally with batch size, the validation losses converged in the latter part of training, signifying that the networks provide similar accuracy. The AP values provided in Table~\ref{tab:sb_tpu} confirm that we are able to train up to a batch size of 2048 with no loss in accuracy. We hypothesize that the increase of 0.6 AP over the original SimpleBaseline implementation (AP of 70.4) was due to training for longer (200 epochs versus 140). Additionally, we demonstrate the importance of warmup and learning rate scaling. When training at the maximum batch size, removing warmup resulted in a loss of 1.3 AP, and removing learning rate scaling resulted in a loss of 0.7 AP.

While preprocessing the data ``online'' on the TPU host CPU provides flexibility for training using different input resolutions and data augmentation, it ultimately causes a bottleneck in the input pipeline. This is evidenced by the training times in Table~\ref{tab:sb_tpu}, which decreased after increasing the batch size to 512, but leveled-off at around 5.3 hours using batch sizes of 512 or greater. We expect that the training time could be reduced substantially if preprocessing and augmentation were included in the TFRecord dataset, or if the TPU host CPU had greater processing capabilities. It is also noted that training these models for 140 epochs instead of 200, as in the original implementation~\cite{xiao2018simple}, reduces the training time to 3.7 hours. Bypassing validation after every epoch speeds up training further. For comparison, training a model of similar size on eight NVIDIA TITAN Xp GPUs takes approximately 1.5 days~\cite{chen2018cascaded}. 

\begin{table}
\footnotesize
\centering
\begin{tabular}{c|c|c|c|c}
	\hline
	\makecell{Batch\\size ($n$)} & Warmup & Scale $l_r$ & \makecell{Training\\Time (hrs)} & AP\\
	\hline
	256 & Y & Y & 7.20 & 71.0\\
	512 & Y & Y & 5.42 & 71.0\\
	1024 & Y & Y & 5.25 & 71.2\\
	2048 & Y & Y & 5.32 & 71.0\\
	\hline
	2048 & N & Y & 5.35 & 69.7\\
	2048 & Y & N & 5.33 & 70.3\\
	\hline
\end{tabular}
\caption{Training times and final AP for large-batch training of SimpleBaseline on Cloud TPU. The original implementation reports an AP of 70.4~\cite{xiao2018simple}. The bottom two rows highlight the importance of warmup and scaling the learning rate when using large batch sizes.}
\label{tab:sb_tpu}
\end{table}

\subsection{Neuroevolution}
\label{sec:neuro_details}
\begin{figure}
\centering
\includegraphics[trim={0, 4mm, 0, 0}, clip, width=\linewidth]{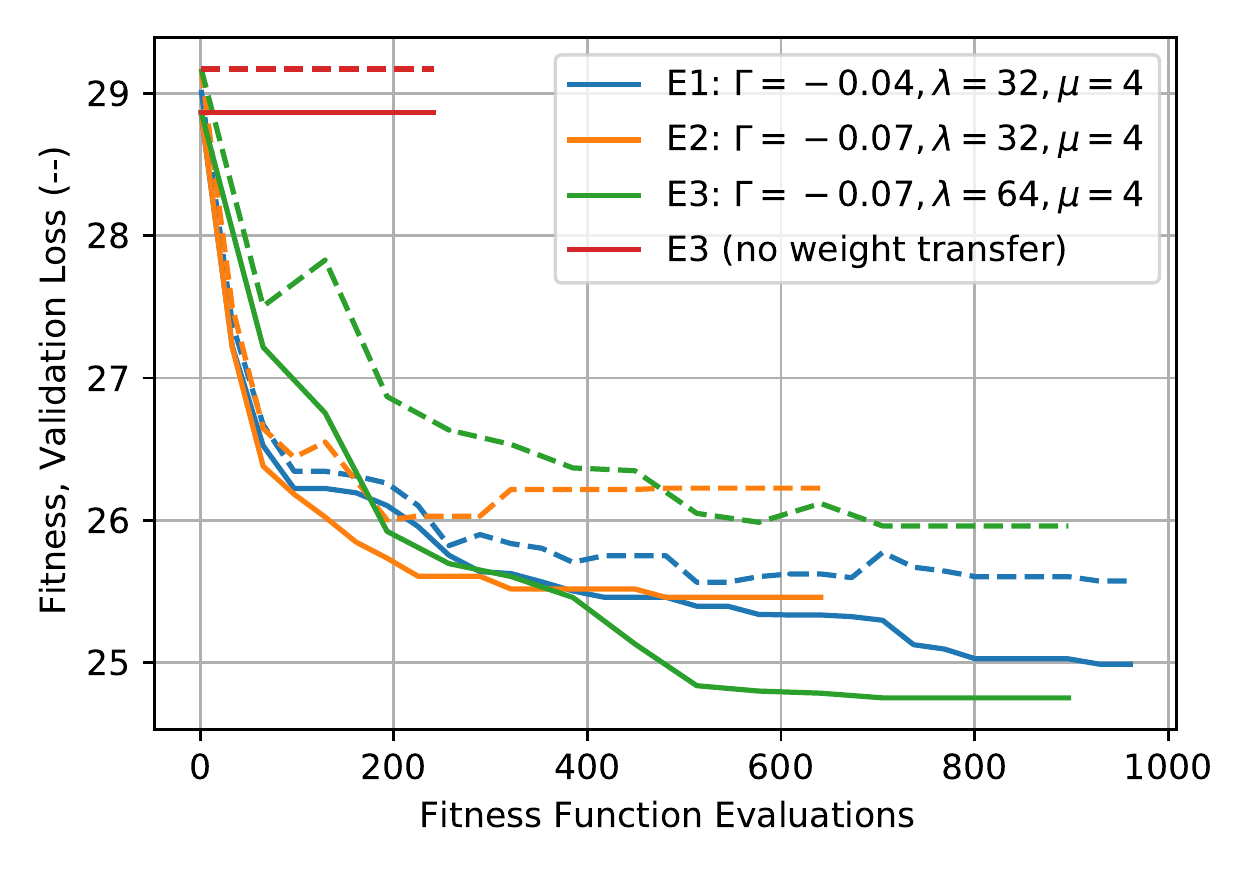}
\includegraphics[trim={2mm, 4mm, 0, 0}, clip, width=0.95\linewidth]{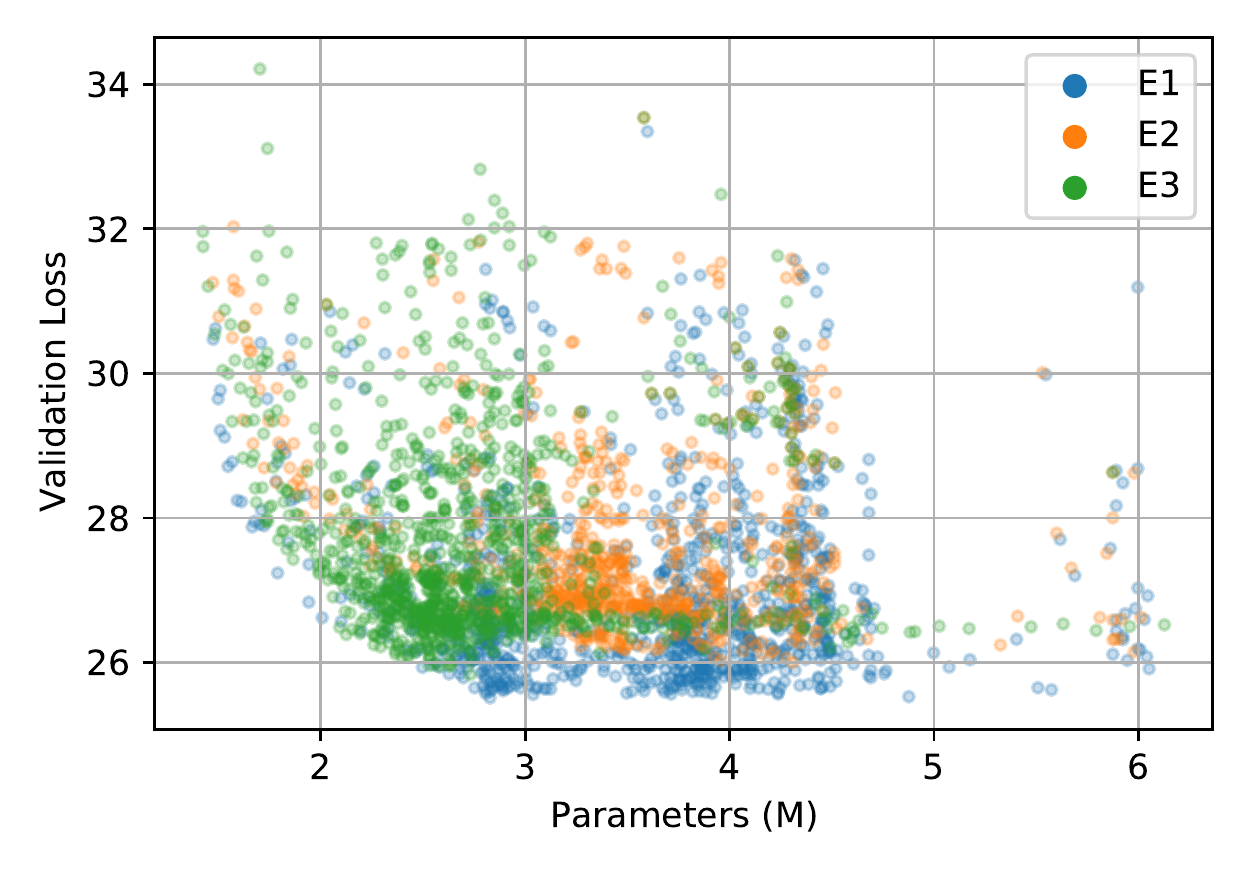}
\caption{Top: Tracking the network with the best fitness in three independent neuroevolutions. The dashed line represents the validation loss of the network with the lowest fitness. $\Gamma$: fitness coefficient controlling trade-off between validation loss and number of parameters. $\lambda$: number of children. $\mu$: number of parents. Bottom: Validation loss versus the number of network parameters for all sampled networks.}
\label{fig:fitness}
\end{figure}

The neuroevolution described in Section~\ref{sec:method} was run under various settings on an 8-CPU, 40 GB memory virtual machine that called on eight v2-8 Cloud TPUs to train several generations of 2D human pose networks. The COCO training and validation sets were used for network training and fitness evaluation, respectively. The input resolution used was 256x192, and the target number of parameters $T$ was set to $5$M. Other settings, including $\Gamma$, $\lambda$, and $\mu$, are provided in the legend of Fig.\ \ref{fig:fitness} (top). ImageNet pretraining was exploited by seeding the common ancestor network using the same inverted residual blocks as used in EfficientNet-B0~\cite{tan2019efficientnet}. The ancestor network was trained for 30 epochs, and we utilize the proposed weight transfer scheme to quickly fine-tune the mutated child networks over just 5 epochs. A batch size of 512 was used to provide near-optimal training efficiency (as per results in previous section) and prevent memory exhaustion mid-search. No learning rate warmup was used during neuroevolution, and the only data augmentation used was horizontal flipping. All other training details are the same as in Section~\ref{sec:tpu_details}.

\subsubsection{Neuroevolution Results}
Fig.\ \ref{fig:fitness} (top) shows the convergence of fitness for three independent neuroevolutions E1, E2, and E3, which had runtimes of 1.5, 0.8 and 1.1 days, respectively. The gap between the fitness (solid line) and validation loss (dashed line) was larger in E2 and E3 compared to E1, indicating that smaller networks were favored more as a result of decreasing $\Gamma$. After increasing the number of children from 32 in E2 to 64 in E3, it became apparent that using fewer children may provide faster convergence, but may also cause the fitness to converge to a local minimum. Fig.\ \ref{fig:fitness} (bottom) plots the validation loss against the number of parameters for all sampled networks. The prominent Pareto frontier near the bottom-left of the figure provides confidence that the search space was thoroughly explored. 

To explicitly demonstrate the benefit of our proposed weight transfer scheme, E3 was run without weight transfer following the same training schedule. As shown in Fig.\ \ref{fig:fitness} (top), the fitness never decreased below that of the ancestor network. It stands that the child networks would need to be trained at least as long as the ancestor network (30 epochs in this case) to achieve the same level of convergence without using the proposed weight transfer scheme. As a result, the neuroevolution runtime would increase six-fold. 

The network with the lowest fitness from neuroevolution E3 was selected as the baseline network, which we refer to as EvoPose2D-S. Its architectural details are provided in Table~\ref{tab:evopose2d-s}. Notably, the overall stride of EvoPose2D-S is less than what is typically seen in hand-designed 2D human pose networks. The lowest spatial resolution observed in the network is $\frac{1}{16}$ the input size, compared to $\frac{1}{32}$ in SimpleBaseline~\cite{xiao2018simple} and HRNet~\cite{sun2019deep}. As a result, the output heatmap is twice as large.

\begin{table}
	\footnotesize
	\centering
	\renewcommand\arraystretch{1.3}
	\begin{tabular}{l|c|c|c|c}
		\hline
		Component & Blocks & \makecell{Kernel\\Size} & Stride & \makecell{Output\\Shape}\\
		\hline
		Stem Conv & - & 3 & 2 & ($\frac{h}{2}$, $\frac{w}{2}$, 32)\\
		\hline
		Module 1 & 1 & 3 & 1 & ($\frac{h}{2}$, $\frac{w}{2}$, 16)\\
		Module 2 & 3 & 3 & 2 & ($\frac{h}{4}$, $\frac{w}{4}$, 24)\\
		Module 3 & 2 & 5 & 2 & ($\frac{h}{8}$, $\frac{w}{8}$, 40)\\
		Module 4 & 4 & 3 & 2 & ($\frac{h}{16}$, $\frac{w}{16}$, 80)\\
		Module 5 & 2 & 5 & 1 & ($\frac{h}{16}$, $\frac{w}{16}$, 112)\\
		Module 6 & 4 & 5 & 1 & ($\frac{h}{16}$, $\frac{w}{16}$, 128)\\
		Module 7 & 2 & 3 & 1 & ($\frac{h}{16}$, $\frac{w}{16}$, 80)\\
		\hline
		Head Conv 1 & - & 3 & 2 & ($\frac{h}{8}$, $\frac{w}{8}$, 128)\\
		Head Conv 2 & - & 3 & 2 & ($\frac{h}{4}$, $\frac{w}{4}$, 128)\\
		Head Conv 3 & - & 3 & 2 & ($\frac{h}{2}$, $\frac{w}{2}$, 128)\\
		\hline
		Final Conv & - & 1 & 1 & ($\frac{h}{2}$, $\frac{w}{2}$, $K$)\\
		\hline
	\end{tabular}
\caption{The architecture of our base 2D human pose network, EvoPose2D-S, designed via neuroevolution. With $h$ = 256, $w$ = 192, and $K = 17$, EvoPose2D-S contains 2.53M parameters and 1.07G FLOPs.}
\label{tab:evopose2d-s}
\end{table}

The baseline network was scaled to various levels of computational expense. A lighter version (EvoPose2D-XS) was created by increasing the stride in Module 6, which halved the number of FLOPs. Using the compound scaling method described in Section~\ref{sec:method}, EvoPose2D-S was scaled to an input resolution of 384x288 (EvoPose2D-M), which is currently the highest resolution used in top-down 2D human pose estimation. We push the boundaries of 2D human pose estimation by scaling to an input resolution of 512x384 (EvoPose2D-L). Even at this high spatial resolution, EvoPose2D-L has roughly half the FLOPs compared to the largest version of HRNet. The scaling parameters for EvoPose2D-M/L are provided in Appendix~\ref{sec:app_scaling}.

\subsubsection{Comparisons with the State-of-the-Art}
\label{sec:sota}
\medskip\noindent\textbf{Microsoft COCO.} To directly compare EvoPose2D with the best methods in the literature, we re-implement SimpleBaseline ResNet-50 (SB-R50) and HRNet-W32 as per the implementation described in Section~\ref{sec:tpu_details}. In our implementation of HRNet, we use a strided transpose pointwise convolution in place of a pointwise convolution followed by nearest-neighbour upsampling. This modification was required to make the model TPU-compatible, and did not change the number of parameters or FLOPs. The accuracy of our implementation is verified against the original in Table~\ref{tab:coco_val}. 

Comparing EvoPose2D-S with our SB-R50 implementation without ImageNet pretraining, we find that EvoPose2D-S provides comparable accuracy on the COCO validation set (see Table~\ref{tab:coco_val}) but is 50\% faster and 12.7x smaller. We also compare EvoPose2D-S to a baseline that stacks the EvoPose2D network head on top of EfficientNet-B0~\cite{tan2019efficientnet}, and find that while EvoPose2D-S is 20\% slower due to its decreased stride, its AP is 1.8 points higher and it is 2.2x smaller. Compared to our HRNet-W32 (256x192) implementation, we observe that EvoPose2D-M is more accurate by 1.5 AP while being 23\% faster and 3.9x smaller. 

\begin{table*}
\footnotesize
\centering
\begin{tabular}{l|c|c|c|c|c|c|c|ccccc}
\hline
Method & PT & Input Size & \makecell{Params\\(M)} & \makecell{FLOPs\\(G)} & \makecell{Size\\(MB)} & \makecell{FPS\\(GPU)} & 
$\operatorname{AP}$ & $\operatorname{AP}^{50}$ & $\operatorname{AP}^{75}$ & $\operatorname{AP}^{M}$ & $\operatorname{AP}^{L}$ & $\operatorname{AR}$ \\
\hline\noalign{\smallskip}
\multicolumn{3}{l}{\textbf{COCO 2017 Validation Dataset}}\\
\hline
CPN~\cite{chen2018cascaded} & Y & $256 \times 192$ & $27.0$ & $6.20$ & $-$ & $-$ &
$68.6$&$-$&$-$&$-$&$-$&$-$\\ 
SB-R50~\cite{xiao2018simple} & Y & $256\times192$  &$34.0$ &$5.21^\dagger$& $137^\dagger$ &  $67.7^\dagger$&
${70.4}$ & ${88.6}$&${78.3}$&${67.1}$&${77.2}$&${76.3}$\\
SB (R-101)~\cite{xiao2018simple} & Y & $256\times192$  &$53.0$ &$8.84^\dagger$& $214^\dagger$ &  $45.1^\dagger$& ${71.4}$ & ${89.3}$&${79.3}$&${68.1}$&${78.1}$&${77.1}$\\
SB (R-152)~\cite{xiao2018simple} & Y & $256\times192$ &$68.6$ &$12.5^\dagger$& $277^\dagger$ &  $34.4^\dagger$&
${72.0}$ & ${89.3}$&${79.8}$&${68.7}$&${78.9}$&${77.8}$\\
HRNet-W32~\cite{sun2019deep} & N & $256\times 192$&  $28.5$ & $7.65^\dagger$ & $119^\dagger$ &  $29.0^\dagger$&
$73.4$&$89.5$&$80.7$&$70.2$&$80.1$&$78.9$  \\
HRNet-W32~\cite{sun2019deep} & Y & $256\times 192$&  $28.5$ & $7.65^\dagger$ & $119^\dagger$ &  $29.0^\dagger$&
$74.4$&$90.5$&$81.9$&$70.8$&$81.0$&$79.8$  \\
HRNet-W48~\cite{sun2019deep}  & Y &  $256\times 192$&  $63.6$ &$15.7^\dagger$ & $259^\dagger$ &  $21.7^\dagger$&
$75.1$&$90.6$&$82.2$&$71.5$&$81.8$&$80.4$  \\
MSPN~\cite{li2019rethinking} & Y &  $256\times 192$&  $120$ &$19.9$ & $-$ & $-$ & $75.9$&$-$&$-$&$-$&$-$&$-$  \\
SB (R-152)~\cite{xiao2018simple} & Y &  $384\times288$ &$68.6$ &$28.1^\dagger$ & $277^\dagger$ &  $24.9^\dagger$
&${74.3}$ & ${89.6}$&${81.1}$&${70.5}$&${79.7}$&${79.7}$\\
HRNet-W32~\cite{sun2019deep} & Y &  $384\times 288$&  $28.5$ &$16.0^\dagger$ & $119^\dagger$ &  $22.7^\dagger$ & $75.8$&$90.6$&${82.7}$&$71.9$&$82.8$&$81.0$  \\
HRNet-W48~\cite{sun2019deep} & Y &  $384\times 288$&  $63.6$ &$35.3^\dagger$ & $259^\dagger$ &  $16.2^\dagger$ & $76.3$&$90.8$&$82.9$&$72.3$&$83.4$&$81.2$  \\
HRNet-W48*~\cite{moon2019posefix} & Y &  $384\times 288$&  $63.6$ &$35.3^\dagger$ & $259^\dagger$ &  $16.2^\dagger$ & $77.3$&$\mathbf{90.9}$&$83.5$&$73.5$&$\mathbf{84.4}$&$82.0$  \\
\hline
SB-R50 & N & $256\times192$  &$34.1$ &$5.21$ & $137$ & $67.7$
&${70.6}$ & ${89.0}$&${78.4}$&${66.9}$&${77.1}$&${77.3}$\\
HRNet-W32 & N & $256\times192$ &$28.6$ &$7.65$ & $119$ &  $29.0$
&${73.6}$ & ${89.9}$&${80.5}$&${70.1}$&${80.0}$&${80.0}$\\
EfficientNet-B0$^\ddagger$ & N & $256\times192$ &${5.82}$ &$0.60$ & $23.9$ &  $123$
&${68.4}$ & ${88.4}$&${76.5}$&${65.0}$&${74.6}$&${75.2}$\\
EvoPose2D-XS & N & $256\times192$ &$\mathbf{2.53}$ &$\mathbf{0.47}$ & $\mathbf{10.8}$ &  $\mathbf{136}$
&${68.0}$ & ${87.9}$&${76.1}$&${64.5}$&${74.3}$&${75.0}$\\
EvoPose2D-S & N & $256\times192$ &$\mathbf{2.53}$ &$1.07$ & $\mathbf{10.8}$ &  $102$
&${70.2}$ & ${88.9}$&${77.8}$&${66.5}$&${76.8}$&${76.9}$\\
EvoPose2D-M & N & $384\times288$ &${7.34}$ &$5.59$ & $30.7$ &  $35.8$
&${75.1}$ & ${90.2}$&${81.9}$&${71.5}$&${81.7}$&${81.0}$\\
EvoPose2D-L & N & $512\times384$ &${14.7}$ &$17.7$ & $60.6$ &  $15.9$
&${76.6}$ & ${90.5}$&${83.0}$&${72.7}$&${83.4}$&${82.3}$\\
EvoPose2D-L* & N & $512\times384$ &${14.7}$ &$17.7$ & $60.6$ &  $15.9$
&$\mathbf{77.5}$ & $\mathbf{90.9}$&$\mathbf{83.6}$&$\mathbf{74.0}$&${84.2}$&$\mathbf{82.5}$\\
\hline\noalign{\smallskip}
\multicolumn{3}{l}{\textbf{COCO 2017 Test Dataset}}\\
\hline
CPN~\cite{chen2018cascaded}& Y & $384\times288$ & - & - & - & - &
$72.1$&$91.4$&$80.0$&$68.7$&$77.2$&$78.5$\\ 
SB (R-152)~\cite{xiao2018simple} & Y & $384\times288$ &$68.6$ &$28.1^\dagger$ & $277^\dagger$ &  $24.9^\dagger$ &
${73.7}$ & ${91.9}$&${81.1}$&${70.3}$&${80.0}$&${79.0}$\\
HRNet-W$48$~\cite{sun2019deep} & Y & $384\times 288$&  $63.6$ &$35.3^\dagger$ & $259^\dagger$ &  $\mathbf{16.2}^\dagger$ & 
$75.5$&${92.5}$&${83.3}$&$71.9$&${81.5}$&$80.5$\\
HRNet-W$48$*~\cite{moon2019posefix} & Y &  $384\times 288$&  $63.6$ &$35.3^\dagger$ & $259^\dagger$ &  $\mathbf{16.2}^\dagger$ & 
$76.7$&$\mathbf{92.6}$&$84.1$&$73.1$&$\mathbf{82.6}$&$81.5$\\
\hline
EvoPose2D-L & N &$512\times384$ &$\mathbf{14.7}$ &$\mathbf{17.7}$ & $\mathbf{60.6}$ &  $15.9$
&${75.7}$ & ${91.9}$&${83.1}$&${72.2}$&${81.5}$&${81.7}$\\
EvoPose2D-L* &  N & $512\times384$ &$\mathbf{14.7}$ &$\mathbf{17.7}$ & $\mathbf{60.6}$ &  $15.9$
&$\mathbf{76.8}$ & ${92.5}$&$\mathbf{84.3}$&$\mathbf{73.5}$&${82.5}$&$\mathbf{81.7}$\\
\hline\noalign{\smallskip}
\multicolumn{8}{l}{\textbf{PoseTrack 2018 Validation Dataset}}\\
\hline
SB-R50 & Y & $256\times192$  &$34.1$ &$5.21$ & $137$ & $67.7$
&${54.3}$ & ${84.8}$&${63.5}$&${30.5}$&${57.7}$&${59.9}$\\
EvoPose2D-S & Y & $256\times192$ &$\mathbf{2.53}$ & $\mathbf{1.07}$ & $\mathbf{10.8}$ &  $\mathbf{102}$
&$\mathbf{55.1}$ & $\mathbf{84.9}$&$\mathbf{64.7}$&$\mathbf{32.4}$&$\mathbf{58.5}$&$\mathbf{60.6}$\\
\hline
HRNet-W32 & Y & $256\times192$ &$28.6$ &$7.65$ & $119$ &  $29.0$
&${58.7}$ & ${87.0}$&${70.2}$&$\mathbf{36.5}$&${62.0}$&${64.3}$\\
EvoPose2D-M & Y & $384\times288$ &${\mathbf{7.34}}$ &$\mathbf{5.59}$ & $\mathbf{30.7}$ &  $\mathbf{35.8}$
&$\mathbf{60.4}$ & $\mathbf{87.9}$&$\mathbf{71.8}$&${36.1}$&$\mathbf{64.0}$&${64.3}$\\
\hline
\end{tabular}
\caption{Comparisons on various datasets. The \texttt{pycocotools} package was used for evaluation. For the COCO datasets, the models in the bottom sections were implemented as per Section~\ref{sec:tpu_details}. For the PoseTrack dataset, all models were implemented as per Section~\ref{sec:sota}. PT: ImageNet pretraining for evaluation on COCO, and COCO pretraining for evaluation on PoseTrack. *: including PoseFix post-processing~\cite{moon2019posefix}. $\ddagger$: EvoPose2D network head stacked on top of EfficientNet-B0. $\dagger$: recalculated for consistency. Network file size provided for float32 models. Frames per second (FPS) averaged over 1k forward passes on NVIDIA TITAN Xp GPU using a batch size of 1. Best results shown in \textbf{bold}.}
\label{tab:coco_val}
\end{table*}


Despite not using ImageNet pretraining, EvoPose2D-L achieves state-of-the-art AP on the COCO validation set\footnote{Higher AP has been reported using HRNet with model-agnostic improvements, including a better person detector and unbiased data processing~\cite{huang2020devil}.} (with and without PoseFix~\cite{moon2019posefix}) while being 4.3x smaller than HRNet-W48. Since EvoPose2D was designed using the COCO validation data, it is especially important to perform evaluation on the COCO test-dev set. We therefore show in Table~\ref{tab:coco_val} that EvoPose2D-L also achieves state-of-the-art accuracy on the test-dev dataset, again without ImageNet pretraining.

\medskip\noindent\textbf{PoseTrack.} For evaluation on PoseTrack, the networks were initialized with the weights pretrained on COCO and were fine-tuned on the Posetrack 2018 training set. All training details are consistent with Section~\ref{sec:tpu_details}, except the fine-tuning process was run for 20 epochs and early-stopping was used. As shown in Table~\ref{tab:coco_val}, the relative performance of EvoPose2D compared to the state-of-the-art is consistent with the COCO dataset: EvoPose2D-S and EvoPose2D-M provide higher accuracy than SB-R50 and HRNet-W32, respectively, despite having fewer parameters and FLOPS, and faster inference speed.

\section{Conclusion}
We propose a simple yet effective weight transfer scheme and use it, in conjunction with large-batch training, to accelerate a neuroevolution of efficient 2D human pose networks. To the best of our knowledge, this is the first application of neuroevolution to 2D human pose estimation. We additionally provide supporting experiments demonstrating that 2D human pose networks can be trained using a batch size of up to 2048 with no loss in accuracy. We exploit large-batch training and the proposed weight transfer to evolve a lightweight 2D human pose network design geared towards mobile deployment. When scaled to higher input resolution, the EvoPose2D network designed using neuroevolution proved to be more accurate than the best performing 2D human pose estimation models in the literature while having a lower computational cost. 

\appendices
\section{Supplementary Material}
\vspace{4pt}
\subsection{Search Space Details}
\label{sec:app_search_space}
A diagram of the hierarchical backbone search space is shown in Fig.\ \ref{fig:search_space}. For each module, we search for the optimal number of blocks, kernel size, output channels, and stride (last three modules only). Table~\ref{tab:ancestor} shows the configuration of the common ancestor network used in our neuroevolution experiments. The kernel size options used were 3x3 and 5x5. The maximum number of blocks was set to four. The maximum number of output channels were set to the values in the common ancestor network (see rightmost column of Table~\ref{tab:ancestor}). Table~\ref{tab:evopose2d-s} shows the architecture of EvoPose2D-S, the network with the best fitness in neuroevolution E3 (see Section~\ref{sec:neuro_details}). The optimal number of output channels in the first five modules were at the upper bound, so it is possible that better results might be obtained if these limits were increased.

\begin{table}[h]
	\footnotesize
	\centering
	\renewcommand\arraystretch{1.3}
	\begin{tabular}{l|c|c|c|c}
		\hline
		Component & \makecell{Blocks\\$N_B$} & \makecell{Kernel\\Size} & Stride & \makecell{Output\\Shape}\\
		\hline
		Module 1 & 1 & 3 & 1 & ($\frac{h}{2}$, $\frac{w}{2}$, 16)\\
		Module 2 & 2 & 3 & 2 & ($\frac{h}{4}$, $\frac{w}{4}$, 24)\\
		Module 3 & 2 & 5 & 2 & ($\frac{h}{8}$, $\frac{w}{8}$, 40)\\
		Module 4 & 3 & 3 & 2 & ($\frac{h}{16}$, $\frac{w}{16}$, 80)\\
		Module 5 & 3 & 5 & 1 & ($\frac{h}{16}$, $\frac{w}{16}$, 112)\\
		Module 6 & 4 & 5 & 2 & ($\frac{h}{32}$, $\frac{w}{32}$, 192)\\
		Module 7 & 1 & 3 & 1 & ($\frac{h}{32}$, $\frac{w}{32}$, 320)\\
		\hline
	\end{tabular}
\caption{Module configuration for the common ancestor network.}
\label{tab:ancestor}
\end{table}

\subsection{Mutation Details}
\label{sec:app_mutation}
The sampled architectures were encoded into 7x4 integer arrays (\# blocks, kernel size, output channels / 8, and stride, for each module), which we refer to as the genotype. The mutations used included increasing/decreasing the number of blocks by 1, changing the kernel size, increasing/decreasing the stride by 1, and increasing/decreasing the number of output channels by 8. During neuroevolution, the genotypes were cached to ensure that no genotype was sampled twice. The mutation function is provided in Algorithm~\ref{algo:mutation}.

\begin{algorithm}
\SetAlgoNoLine
\small
\DontPrintSemicolon
\newcommand\mycommfont[1]{\footnotesize\ttfamily\textcolor{blue}{#1}}
\SetCommentSty{mycommfont}

\SetKwInput{KwInput}{Input}                
\SetKwInput{KwOutput}{Output}              
\KwInput{parent genotype $g_p$, ancestor genotype $g_a$, genotype cache $G$}
\KwOutput{mutated child genotype $g_c$}
 $g_c \leftarrow g_p$\;
 \While{$g_c$ \normalfont\texttt{in}  $ G $  \normalfont\texttt{or} $g_c = g_p$}{
     $g_c \leftarrow g_p$\;
     $i, j \leftarrow \texttt{randint}(7), \texttt{randint}(4)$\;
     \If{$j = 0$}{
        \tcp{mutate number of blocks}
        \If{$g_c[i, j] = 1$}{$g_c[i, j] \mathrel{+}= 1$\;}
        \ElseIf{$g_c[i, j] = 4$}{$g_c[i, j] \mathrel{-}= 1$\;}
        \ElseIf{$ \normalfont\texttt{randint}(2) > 0$}
            {$g_c[i, j] \mathrel{+}= 1$\;}
        \Else{$g_c[i, j] \mathrel{-}= 1$\;}
     }
     \ElseIf{$j = 1$}{
     \tcp{mutate kernel}
      $g_c[i, j] \leftarrow \{3, 5\}[\texttt{randint}(2)]$ \;}
     \ElseIf{$j = 2$}{
      \tcp{mutate output channels}
      $g_c[i, j] \leftarrow \texttt{randint}(g_a[i, j]) + 1$}
     \ElseIf{$j = 3$ \normalfont\texttt{and} $i \geq 4$}{
      \tcp{mutate stride}
      \If{$g_c[i, j] = 2$ \normalfont\texttt{and} \normalfont\texttt{sum}$(g_p[:, j] - 1) = 4$}{$g_c[i, j] \mathrel{-}= 1$\;}
      \ElseIf{$g_c[i, j] = 1$ \normalfont\texttt{and} \normalfont\texttt{sum}$(g_p[:, j] - 1) < 4$}{$g_c[i, j] \mathrel{+}= 1$\;}
  }
 } $G$\texttt{.append}($g_c$)

 \caption{Mutation}\label{algo:mutation}
\end{algorithm}

\subsection{Weight Transfer Details}
\label{sec:app_wt}
The child network architectures were decoded from the mutated child genotypes, and all weights in the child networks were randomly initialized. Then, the weight transfer scheme described in Section~\ref{sec:weight_transfer} was used to transfer the trained weights from the parents to the children. For batch normalization layers, the non-transferred weights were initialized with the means of the parent. When a new block was added as a result of a mutation, the weights from the parent's last block were transferred to the new block in the child.

\subsection{Scaling Coefficients}
\label{sec:app_scaling}
The scaling coefficients used for EvoPose2D-M and EvoPose2D-L are provided in Table~\ref{tab:scaling_coefficients}. $c_d$ scales the number of blocks in each module, rounded to the nearest integer. $c_w$ scales the number of output channels used in each block, rounded to the nearest multiple of eight. See Eq.\ \ref{eq:scaling} for details on how these values were calculated. 

\begin{table}[h]
\footnotesize
\centering
\renewcommand\arraystretch{1.3}
\begin{tabular}{l|c|c|c|c}
	\hline
	Model & Input Size & $\phi$ & $c_d$ & $c_w$\\
	\hline
	EvoPose2D-M & $384\times 288$ & 2.90 & 1.70 & 1.32\\
	EvoPose2D-L & $512\times 384$ & 4.96 & 2.47 & 1.60\\
	\hline
\end{tabular}
\caption{Scaling coefficients for EvoPose2D-M/L}
\label{tab:scaling_coefficients}
\end{table}

\bibliographystyle{IEEEtran}
\bibliography{egbib}

\begin{IEEEbiography}[{\includegraphics[width=1in,height=1.25in,clip,keepaspectratio]{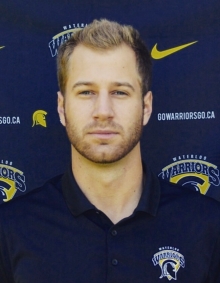}}]{William McNally} is a Ph.D. Candidate in the Department of Systems Design Engineering at the University of Waterloo, Canada. He received the B.Eng. in Mechanical Engineering from Memorial University of Newfoundland, Canada in 2015, and the M.A.Sc. in Systems Design Engineering from the University of Waterloo in 2018. He has authored 18 peer-reviewed journal and conference papers in research areas ranging from biomechanics and sports engineering to computer vision and deep learning. He is the recipient of a number of scholarships and awards, including the President's Graduate Scholarship, the Ontario Graduate Scholarship, the Alexander Graham Bell Canada Graduate Scholarship, the JS Rancourt Golf Excellence Award, the Best Overall Paper award at the 2018 Conference on Computer Vision and Imaging Systems, and the Best Paper award at the ICML 2021 Workshop on Tackling Climate Change with Machine Learning. 
\end{IEEEbiography}

\begin{IEEEbiography}[{\includegraphics[width=1in,height=1.25in,clip,keepaspectratio]{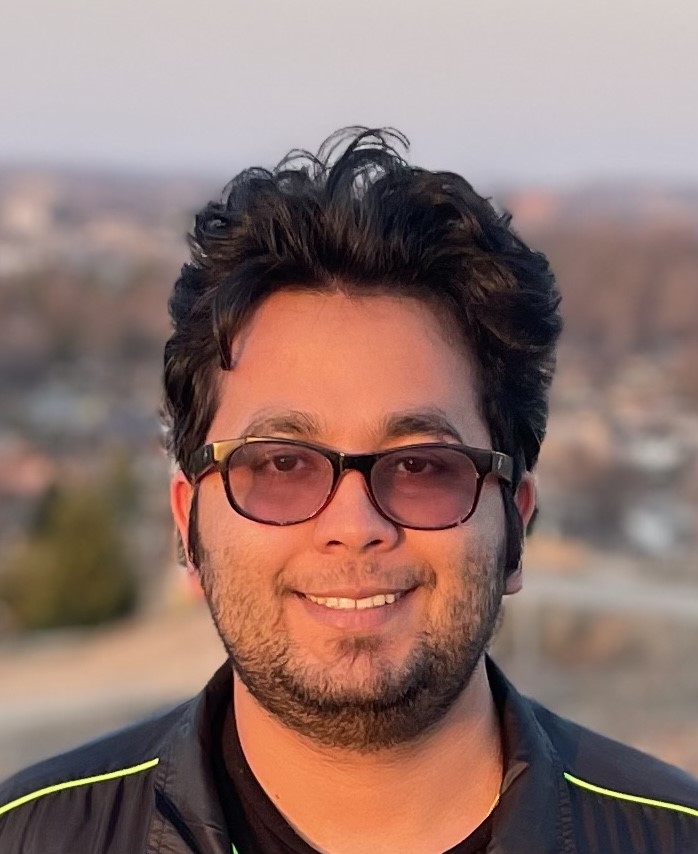}}]{Kanav Vats} is a Ph.D. student in the Department of Systems Design Engineering at the University of Waterloo, Canada. He received his Integrated Masters of Science in Applied Mathematics at the Indian Institute of Technology Roorkee in 2018. He has authored 12 peer-reviewed research papers in computer vision and deep learning. His research interest lies in applied computer vision-based sports analytics, with a focus on problems such as player tracking, player identification, and event detection from broadcast video.
\end{IEEEbiography}

\begin{IEEEbiography}[{\includegraphics[width=1in,height=1.25in,clip,keepaspectratio]{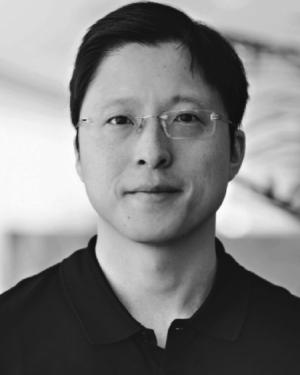}}]{Alexander Wong} (M’06–SM’21) received the B.A.Sc. degree in computer engineering, the M.A.Sc. degree in electrical and computer engineering, and the Ph.D. degree in Systems Design Engineering from the University of Waterloo, Waterloo, ON, Canada, in 2005, 2007, and 2010, respectively. He is currently the Canada Research Chair of Artificial Intelligence and Medical Imaging, the Co-Director of the Vision and Image Processing Research Group, an Associate Professor with the Department of Systems Design Engineering, University of Waterloo, and a member of the College of the Royal Society of Canada. 

He has authored over 560 refereed journal and conference papers and patents, in various fields, such as computational imaging, artificial intelligence, computer vision, graphics, image processing, and multimedia systems. His research interests focus on integrative biomedical imaging systems design, operational artificial intelligence, and scalable and explainable deep learning. He has received a number of awards, including two Outstanding Performance Awards, the Distinguished Performance Award, the Engineering Research Excellence Award, the Sandford Fleming Teaching Excellence Award, the Early Researcher Award from the Ministry of Economic Development and Innovation, the Outstanding Paper Award at the CVPR Workshop on Adversarial Machine Learning in Real-World Computer Vision Systems and Online Challenges (2021), the Best Paper Award at the NIPS Workshop on Transparent and Interpretable Machine Learning (2017), the Best Paper Award at the NIPS Workshop on Efficient Methods for Deep Neural Networks (2016), the two Best Paper Awards by the Canadian Image Processing and Pattern Recognition Society in 2009 and 2014, respectively, the Distinguished Paper Award by the Society of Information Display (2015), the two Best Paper Awards for the Conference of Computer Vision and Imaging Systems in 2015 and 2017, respectively, the Synaptive Best Medical Imaging Paper Award (2016), the two Magna Cum Laude Awards and one Cum Laude Award from the Annual Meeting of the Imaging Network of Ontario, the AquaHacking Challenge First Prize (2017), the Best Student Paper at Ottawa Hockey Analytics Conference (2017), and the Alumni Gold Medal.
\end{IEEEbiography}

\begin{IEEEbiography}[{\includegraphics[width=1in,height=1.25in,clip,keepaspectratio]{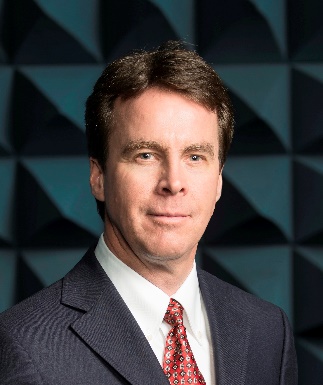}}]{Professor John McPhee} is the Canada Research Chair in System Dynamics at the University of Waterloo, Canada, which he joined in 1992. Prior to that, he held fellowships at Queen’s University, Canada, and the Université de Liège, Belgium.  

He pioneered the use of linear graph theory and symbolic computing to create real-time models and model-based controllers for multi-domain dynamic systems, with applications ranging from autonomous vehicles to rehabilitation robots and sports engineering. His research algorithms are a core component of the widely-used MapleSim modelling software, and his work appears in more than 160 journal publications. 

Prof. McPhee is the past Chair of the International Association for Multibody System Dynamics, a co-founder of 2 international journals and 3 technical committees, a member of the Golf Digest Technical Panel, and an Associate Editor for 5 journals. He is a Fellow of the Canadian Academy of Engineering, the American and Canadian Societies of Mechanical Engineers, and the Engineering Institute of Canada.  He has won 8 Best Paper Awards and, in 2014, he received the prestigious NSERC Synergy Award from the Governor-General of Canada. 
\end{IEEEbiography}

\EOD

\end{document}